\documentclass[conference, review]{IEEEtran}

\usepackage{cite}
\usepackage{amsmath,amssymb,amsfonts}
\usepackage{algorithmic}
\usepackage{graphicx}
\usepackage{enumitem}
\usepackage{listings}
\usepackage{orcidlink}

\usepackage{hyperref}

\usepackage{algorithm}    
\usepackage{algorithmic}  

\usepackage{textcomp}
\usepackage{xcolor}
\def\BibTeX{{\rm B\kern-.05em{\sc i\kern-.025em b}\kern-.08em
    T\kern-.1667em\lower.7ex\hbox{E}\kern-.125emX}}
\begin{document}

\title{Krony-PT: GPT-2 Compressed with Kronecker Products \\
}


\author{
\IEEEauthorblockN{Mohamed Ayoub Ben Ayad \orcidlink{0009-0009-0774-5412}}
\IEEEauthorblockA{\textit{Chair of Data Science} \\
\textit{University of Passau}\\
Passau, Germany\\
Mohamed.Benayad@uni-passau.de}
\and
\IEEEauthorblockN{Jelena Mitrovi\'{c}
\orcidlink{0000-0003-3220-8749}}
\IEEEauthorblockA{\textit{Chair of Data Science} \\
\textit{University of Passau}\\
Passau, Germany\\
Jelena.Mitrovic@uni-passau.de}
\and
\IEEEauthorblockN{Michael Granitzer
\orcidlink{0000-0003-3566-5507}}
\IEEEauthorblockA{\textit{Chair of Data Science} \\
\textit{University of Passau}\\
Passau, Germany\\
Michael.Granitzer@uni-passau.de}
}

\maketitle
\IEEEpeerreviewmaketitle

\begin{abstract}
We introduce Krony-PT, a compression technique for GPT-2 based on Kronecker products. We specifically target the feed-forward weights of each transformer block, and systematically compress the feed-forward layer matrices to various degrees. We introduce a modified Van Loan decomposition to initialize new Kronecker factors, and also propose a new pruning-based initialization technique. Our method compresses the original 124M-parameter GPT-2 to various smaller models, ranging from 80M to 96M. Our 81M model variant outperforms DistilGPT2 on next-token prediction across all standard language modeling datasets, and shows competitive or comparable performance with significantly larger Kronecker-based compressions of GPT-2.

\end{abstract}


\section{Introduction}
Given their rapid development, Large Language Models (LLMs) have revolutionized numerous industries and changed the way we perceive and interact with machines. Their ability to understand, generate, and generalize across various domains, including automated text generation, language translation, and notably code synthesis, has led many to believe in the scaling hypothesis \cite{branwen2021scaling}, the idea that scaling model size is ``all you need'' to achieve Artificial General Intelligence (AGI). Despite their advanced features, the core of LLMs still fundamentally relies on the original Transformer architecture \cite{vaswani2017attention}, particularly the decoder-only variants popularized by the release of GPT-2 \cite{radford2019language}. 

While large corporations can afford the vast computational resources and data required to train state-of-the-art models like GPT-4 \cite{achiam2023gpt} and its competitors \cite{claude3, llama3, reid2024gemini}, individuals and smaller organizations face significant barriers to reproduce these frontier models. Therefore, advancing research on LLM compression and factorization is essential to make these powerful tools more accessible, sustainable, and deployable on resource-constrained hardware, such as edge devices.


Matrix Factorization techniques leverage ideas from classical linear algebra (e.g., SVD, QR, and Kronecker products) to reduce the size and parameter count of LLMs. 
In particular, the Kronecker product \cite{schacke2004kronecker} 
offers an effective method for decomposing large matrices into structured, smaller blocks. 
For instance, using the Van Loan decomposition (detailed in Section~\ref{sec:VL}), a singular value decomposition (SVD)-based method, we can optimally\footnote{In the Frobenius norm sense.} express any given matrix as a sum of multiple Kronecker products of smaller matrices. Kronecker product decompositions (with appropriate factor dimensions, see Section \ref{sec:rank}) can achieve a significant parameter reduction while preserving the rank of the original matrix, in contrast to methods that specifically aim for a low-rank approximation \cite{ben-noach-goldberg-2020-compressing, hsu2022language}. Kronecker products also provide a natural framework for vectorizing operations and accelerating computation.

This paper builds on these methods by applying Kronecker-based factorizations to the MLP layers of GPT-2. We extend the existing literature on Kronecker-based compression of LLMs by introducing various rank-invariant compression schemes, with key architectural changes to the MLPs of each transformer layer, leading to models of different sizes. We also introduce two lightweight initialization techniques that significantly improve and stabilize convergence. 

\textbf{Our contributions}\footnote{The implementation is available at: \url{https://github.com/eigenAyoub/krony-PT/}} could be summarized as follows:
\begin{itemize}
    \item We explore different compression schemes that produce models of various sizes, including 81M, 92M, and 96M. 
    \item We propose a new pruning-based initialization for the 96M model, and an improved scale-invariant Van Loan (VL) decomposition for the other models.
    \item We enhance the decomposition by introducing learnable scalars for each Kronecker factor, providing an additional degree of freedom without a significant increase in parameters or inference overhead.
    \item We apply a consistent compression strategy across all transformer layers, ensuring a uniform architectural modification, unlike prior work that selectively compresses layers.
    \item We use weight tying for the embedding and final output matrices. This practice, standard for models like GPT-2, reduces the parameter count by 38M. Other Kronecker-based methods \cite{edalati2021kroneckr,abronin2024tqcompressor} do not employ this, resulting in parameter counts that are not directly comparable to ours (see Appendix \ref{sec:param-count} for details).
\end{itemize}

\section{Related Work}

There are three major lines of research in the compression of LLMs: \textbf{Distillation} 
\cite{hinton2015distilling, gou2021knowledge, sanh2019distilbert}, 
\textbf{Quantization} \cite{chen2020statistical, lin2020towards, dettmers2022gpt3},
and \textbf{Factorization}
\cite{noach2020compressing}.
Our work focuses on the latter category, Matrix Factorization provides a straightforward framework to compress models. Generally speaking, Matrix Factorization techniques aim to reconstruct an original (typically larger) matrix by combining smaller, often lower-rank, matrices (e.g., via a product). With techniques ranging from SVD \cite{golub1971singular, sainath2013low}, to other low-rank factorization methods \cite{xue2013restructuring}, while the use of Kronecker products remains less common.

Kronecker products \cite{van1993approximation, graham2018kronecker} provide a powerful framework for matrix manipulation and can simplify complex computations through more compact and intuitive notation (e.g., computing the Kronecker-based Fisher matrix approximations \cite{martens2015optimizing}). Their use naturally arises in machine learning, where they help model and accelerate computations in large neural networks, and reduce the notational burden, as demonstrated in various applications \cite{leskovec2010kronecker, gao2020kronecker}. Additionally, Kronecker products can also be used to approximate a given matrix by two smaller matrices, e.g., using 
decomposition methods like Van Loan's \cite{van1993approximation}, hence greatly reducing the total number of parameters of the model.

Using Kronecker products as a factorization technique first appeared in the literature to approximate the weight matrices in fully connected neural networks by a sum of Kronecker products \cite{wu2016compression}, and was then used to compress LLMs. This approach was most notable for encoder-only models like BERT \cite{tahaei2022kroneckerbert}, and more recently applied to decoder-only methods like GPT-2 \cite{edalati2021kroneckr,abronin2024tqcompressor}.

\section{Methodology}

This section details our compression methodology for the GPT-2 Small (124M) model, which we refer to as GPT-2 for the remainder of this paper. We describe our factorization approach, define the resulting compression schemes, and highlight key design choices that distinguish our work from prior art.

\subsection{Compression Strategy and Key Differentiators}

Our compression strategy targets the most parameter-dense components of the GPT-2 architecture. A standard GPT-2 model consists of 12 transformer layers, each containing a self-attention module and a two-layer feed-forward network (FFN). We exclusively compress the weight matrices of these FFNs\footnote{Referred to as \texttt{c\_fc.weight} and \texttt{c\_proj.weight} in the code.}. Collectively, these matrices account for 56.6M parameters, amounting to 45\% of the model's total 124M parameters. Compressing them therefore offers a path to significant model size reduction. Our methodology differs from prior Kronecker-based compression work \cite{edalati2021kroneckr,abronin2024tqcompressor} in two fundamental ways:
\begin{itemize}
    \item \textbf{Uniform Layer Compression:} We apply our factorization scheme identically to all 12 transformer layers, creating a homogeneous architecture, in contrast to methods that only compress odd-numbered layers.
    \item \textbf{Weight Tying:} We employ weight tying for the input embedding and final output projection matrices, a standard practice that reduces the parameter count by 38M. This is a critical difference that makes direct parameter-count comparisons more nuanced (see Appendix \ref{sec:param-count}).
\end{itemize}

\subsection{An Introductory Example: The 96M Model}
\label{sec:96M}

The most direct factorization strategy is one that effectively halves one dimension of each of the two large FFN weight matrices. This approach leads to our 96M-parameter model. 

Let the two FFN matrices for a given layer be $p_1 \in \mathbb{R}^{3072 \times 768}$ and $p_2 \in \mathbb{R}^{768 \times 3072}$, which we decompose using a Kronecker product in the following fashion: 
\[
p_1 = \underbrace{W_{11}}_{\text{(3072,384)}} \otimes \underbrace{W_{12}}_{\text{(1,2)}}
\quad \& \quad
p_2 = \underbrace{W_{21}}_{\text{(384,3072)}} \otimes \underbrace{W_{22}}_{\text{(2,1)}}
\]
This factorization reduces the FFN parameters by 50\% (a 28M reduction), resulting in a final model of approximately 96M parameters. While this is a substantial compression, our ultimate goal is to achieve an 81M model for a fair and direct comparison with DistilGPT-2 \cite{sanh2019distilbert}. To reach this more aggressive compression target, we must move beyond this simple dimensional split and explore a broader design space of factorization schemes.

\subsection{The Kronecker Factorization Design Space}
\label{sec:rank}

The 96M model represents a single point in a design space of possible decompositions. This space is governed by a fundamental trade-off: minimizing the parameter count versus preserving the expressive capacity of the original matrix. A key measure of this capacity is matrix rank. For Kronecker products, rank is multiplicative: $\mathrm{rank}(A\otimes B) = \mathrm{rank}(A)\cdot\mathrm{rank}(B)$.\footnote{For a proof, see: \url{https://math.stackexchange.com/questions/707091/elementary-proof-for-textrank-lefta-otimes-b-right-textranka-cdot}}

Formally, for an FFN matrix $W \in \mathbb{R}^{m \times n}$ (where $m=3072, n=768$), we seek an approximation $W \approx A \otimes B$ with factors $A \in \mathbb{R}^{m_1 \times n_1}$ and $B \in \mathbb{R}^{m_2 \times n_2}$. The dimensions must satisfy $m = m_1 m_2$ and $n = n_1 n_2$. The number of parameters in this factorized form is $m_1 n_1 + m_2 n_2$. Our goal is to find factor dimensions that significantly reduce this parameter count while preserving the maximal rank of the original matrix, which is $\min(m, n) = 768$.

Table~\ref{tab:compression-schemes} summarizes several decomposition schemes that satisfy this maximal-rank criterion, detailing their resulting FFN parameter counts and total model sizes. Our experimental investigation focuses primarily on the models in the 96M, 81M.

Furthermore, the single-factor decomposition ($W \approx A \otimes B$) can be extended to a sum of multiple factors ($W \approx \sum_{i=1}^k A_i \otimes B_i$). This provides another degree of freedom to control the parameter-performance trade-off. Table~\ref{tab:multi-kronecker-factors} shows how the model size increases as we add more factors for specific base dimensions. Our work also explores a 4-factor model which results in a total size of 92M parameters.

\begin{table}[!t]
  \renewcommand{\arraystretch}{1.2}  
  \caption{Rank-preserving Kronecker decomposition schemes for FFN layers}
  \label{tab:compression-schemes}
  \centering
  \begin{tabular}{|l|c|c|c|}
    \hline
    \textbf{Name}    & \textbf{Dimension}   & \textbf{\# Params} & \textbf{Model size} \\ 
    \hline
    \textbf{67M}     & (64, 32)   & 3200    & 67,893,504  \\
                    & (64, 48)   & 3840    & 67,908,864  \\
                    & (96, 32)   & 3840    & 67,908,864  \\
                    & (64, 64)   & 4672    & 67,928,832  \\
                    & (128, 32)  & 4672    & 67,928,832  \\
                    & (96, 48)   & 5120    & 67,939,584  \\
                    & (96, 64)   & 6528    & 67,973,376  \\
                    & (128, 48)  & 6528    & 67,973,376  \\
    \hline
    \textbf{68M}     & (128, 64)  & 8480    & 68,020,224  \\
                    & (96, 96)   & 9472    & 68,044,032  \\
                    & (192, 48)  & 9472    & 68,044,032  \\
                    & (128, 96)  & 12480   & 68,116,224  \\
                    & (192, 64)  & 12480   & 68,116,224  \\
    \textbf{MF1}    & (128, 128) & 16528   & 68,213,376  \\
                    & \ldots     & \ldots  & \ldots      \\
    \textbf{MF2}    & (1024, 256)& 262153  & 74,108,376  \\
                    & (768, 384) & 294920  & 74,894,784  \\
                    & (1024, 384)& 393222  & 77,254,032  \\
    \hline
    \textbf{81M}     & (768, 768)   & 589828  & 81,972,576  \\
                    & (1536, 384)  & 589828  & 81,972,576  \\
                    & (1024, 768)  & 786435  & 86,691,144  \\
    \textbf{96M}     & (1536, 768)  & 1179650 & 96,128,304  \\
    \hline
    \textbf{GPT-2}    & (3072, 768) & 2359297 & 124,439,832 \\
    \hline
  \end{tabular}
\end{table}

\begin{table}[!t]
  \renewcommand{\arraystretch}{1.2}  
  \caption{Adding multiple Kronecker factors}
  \label{tab:multi-kronecker-factors}
  \centering
  \begin{tabular}{|l|c|c|c|c|c|}
    \hline
    \textbf{Name} & \textbf{Dimension} & \textbf{\#params} & \textbf{1 factor} & \textbf{2 factors} & \textbf{3 factors} \\
    \hline
    MF1 & (256, 64)   & 16\,528  & 68.2\,M & 68.6\,M & 69\,M \\
    MF2 & (1024, 256) & 262\,153 & 74\,M   & 80\,M   & 86\,M \\
    \hline
  \end{tabular}
\end{table}

\subsection{Initialization Strategies}

Having defined the target architecture of the compressed model, the crucial next step is to initialize the new factor weights. Since we start from a pre-trained GPT-2 checkpoint, our goal is to initialize the factors in a way that preserves as much knowledge from the original pre-trained model as possible. While parameters common to both models can be copied directly, initializing the new Kronecker factors is more challenging. 

In this work, we propose and evaluate two distinct initialization strategies:
\begin{enumerate}[label=(\roman*)]
    \item A novel rescaling of the standard Van Loan (VL) decomposition that preserves the norm of the original weight matrix, detailed in Section~\ref{sec:norm-init}.
    \item A new pruning-based initialization that forgoes SVD entirely, described in Section~\ref{sec:prun-init}.
\end{enumerate}
Both approaches are compared against the standard VL decomposition, which we describe first as a baseline.

\subsection{Baseline: The Van Loan Decomposition}
\label{sec:VL}

The standard method for initializing Kronecker factors from an original matrix is the Van Loan (VL) decomposition \cite{van1993approximation}. This SVD-based algorithm finds a set of factor pairs $\{U_i, V_i\}$ whose Kronecker product sum, $\sum_{i=1}^k U_i \otimes V_i$, is the optimal rank-$k$ approximation of a target matrix $W$ in the Frobenius norm sense ($\arg\min \left\| W - \sum_{i=1}^{k} U_i \otimes V_i \right\|_F$). 
The algorithm (detailed in Algorithm~\ref{algo:VL}) essentially rearranges the target matrix $W$ to isolate the structure of the factors, performs an SVD to find the most significant components, and then reshapes these components back into the desired factor dimensions. A Python implementation is provided in Appendix~\ref{appendix:kronecker}.

\begin{algorithm}[!t]
  \caption{Kronecker Product Decomposition}
  \label{algo:VL}
  \begin{algorithmic}[1]
    \REQUIRE Matrix $A$, integers $m$, $m_2$, $n$, $n_2$, and rank $k$
    \ENSURE  Kronecker factors $\{U_i\}$ and $\{V_i\}$
    \STATE Rearrange $A$ into a matrix suitable for Kronecker decomposition
    \STATE Compute thin SVD:
      \[
        A \approx U\,S\,V^T
      \]
      retaining the top $k$ singular values
    \STATE Reshape and scale the columns of $U$ and $V$ to the desired factor dimensions
    \STATE \textbf{return} $\{U_i\}$ and $\{V_i\}$ scaled by $\sqrt{S}$
  \end{algorithmic}
\end{algorithm}

\paragraph{Multiple factors with scalars.} Algorithm ~\ref{algo:VL} returns $k$ factors.  In our work, we take advantage of each Kronecker factor $A_i \otimes B_i$ and add it with an additional degree of freedom in the form of a scalar $s_i$, resulting in our MLP weight being represented as follows $W  = \sum_{n=1}^{k} s_i (A_i \otimes B_i)$, Table \ref{tab:dimension-table} details two multi-factor configurations used in our experiments (see Sections \ref{sec:81M} and \ref{multi-kron}).

\begin{table}[!t]
  \renewcommand{\arraystretch}{1.2}  
  \caption{Comparison of dimensions, factors, parameters, and model names}
  \label{tab:dimension-table}
  \centering
  \begin{tabular}{|l|c|c|c|c|}
    \hline
    \textbf{Model name} & \textbf{$A_i$ dim} & \textbf{$B_i$ dim} & \textbf{\# factors} & \textbf{\# parameters} \\
    \hline
    KronyPT-80M-2 & (1024, 256) & (3, 3) & 2 & 80\,M   \\
    KronyPT-92M-4 & (1024, 256) & (3, 3) & 4 & 92.2\,M \\
    \hline
  \end{tabular}
\end{table}

While the VL framework is mathematically optimal for reconstruction, we identified a key limitation of this approach when applied to pre-trained neural networks.

\subsection{Method 1: Adaptive Normalization for VL}
\label{sec:norm-init}

A critical issue with the standard VL decomposition is its failure to preserve the norm of the original weight matrices. This can disrupt the carefully balanced signal magnitudes of the original pre-trained model. Table~\ref{tab:model-comparison} quantifies this problem, showing that the Frobenius norm of the factorized weights can be as low as 67\% of the original's, a significant drop in magnitude.

\begin{table}[!t]
  \renewcommand{\arraystretch}{1.2}
  \caption{Comparison of norms of some parameters across different layers for three models: GPT-2, Krony-PT-96M, and Krony-PT-92M.}
  \label{tab:model-comparison}
  \centering
  \begin{tabular}{|l|l|l|c|c|c|c|}
    \hline
    \textbf{Layer}   & \textbf{Param.} & \textbf{Norm} 
      & \textbf{GPT-2}
      & \textbf{96M}
      & \textbf{92M}
      & \textbf{\%} \\
    \hline
    Layer~1          & $p_1^1$            & Frobenius   & 216   & 153   & 146   & 67.6\% \\
                     &                    & $\ell_1$    & 248~k & 154~k & 161~k &        \\
    \cline{2-7}
                     & $p_2^1$            & Frobenius   & 135   & 95    & 91    & 67.4\% \\
                     &                    & $\ell_1$    & 151~k & 84~k  & 94~k  &        \\
    \hline
    Layer~2          & $p_1^2$            & Frobenius   & 200   & 142   & 134   & 67\%   \\
                     &                    & $\ell_1$    & 235~k & 132~k & 157~k &        \\
    \cline{2-7}
                     & $p_2^2$            & Frobenius   & 133   & 94    & 92    & 69.1\% \\
                     &                    & $\ell_1$    & 151~k & 103~k & 91~k  &        \\
    \hline
    Layer~11         & $p_1^{11}$         & Frobenius   & 199   & 141   & 134   & 67.3\% \\
                     &                    & $\ell_1$    & 241~k & 170~k & 161~k &        \\
    \cline{2-7}
                     & $p_2^{11}$         & Frobenius   & 304   & 217   & 204   & 67.1\% \\
                     &                    & $\ell_1$    & 335~k & 243~k & 223~k &        \\
    \hline
  \end{tabular}
\end{table}

To address this, we introduce a simple but effective adaptive normalization step. Let $W$ be the original weight matrix from GPT-2 and $\hat{W} = \sum_{i=1}^k A_i \otimes B_i$ be its standard VL approximation up to $k$ factors. We compute a rescaling factor $\alpha_{W}$ as the ratio of their norms:
\[
\alpha_{W} = \frac{\| W \|_F}{\| \hat{W} \|_F}
\]
We then define our initialized matrix, $\hat{W}_{\text{norm}}$, as the rescaled approximation:
\[
\hat{W}_{\text{norm}} = \alpha_{W} \hat{W}
\]
This ensures our initialized weights have the exact same Frobenius norm as the original weights, i.e., $\| \hat{W}_{\text{norm}} \|_F = \| W \|_F$. When using multiple factors with learnable scalars ($W = \sum s_i(A_i \otimes B_i)$), we simply initialize each scalar $s_i$ to $\alpha_W$. Except for the 96M model using pruning (detailed in the next section), all of our models are initialized with this adaptive normalization technique.

\subsection{Method 2: Pruning-Based Initialization}
\label{sec:prun-init}

Prior work has shown that optimal, task-agnostic weight approximations do not necessarily translate to optimal performance in compressed neural networks \cite{hsu2022language}. This is mostly because these methods aim to minimize reconstruction error independently of the downstream task.

Motivated by these potential limitations, we propose an alternative initialization strategy that completely avoids the SVD-based VL decomposition. 
Our method induces sparsity in the first factor of the Kronecker product, which is equivalent (for the 96M model variant) to pruning the original matrix by half. This is equivalent to setting one of the Kronecker factors to a fixed vector like $[1, 0]^T$, as illustrated in Figure~\ref{fig:small-pruning-illustration}.

To ensure stable training and avoid the zero initialization issue, we instead use $[1, 0.1]^T$ for the second factor. 
This straightforward, SVD-free method provides a serious alternative to the standard VL initialization used in prior work, and we compare their performance in Table~\ref{tab:prune-results}.

\begin{figure}[!t]
  \centering
  \resizebox{.9\columnwidth}{!}{%
    $\displaystyle
      \begin{bmatrix}
        a_{11} & a_{12} & a_{13} & a_{14} \\
        a_{21} & a_{22} & a_{23} & a_{24} \\
        a_{31} & a_{32} & a_{33} & a_{34} \\
        a_{41} & a_{42} & a_{43} & a_{44} \\
      \end{bmatrix}
      \xrightarrow[\text{pruning}]{}
      \begin{bmatrix}
        a_{11} & a_{12} & a_{13} & a_{14} \\
        0      & 0      & 0      & 0      \\
        a_{31} & a_{32} & a_{33} & a_{34} \\
        0      & 0      & 0      & 0      \\
      \end{bmatrix}
      =
      \begin{bmatrix}
        a_{11} & a_{12} & a_{13} & a_{14} \\
        a_{31} & a_{32} & a_{33} & a_{34} \\
      \end{bmatrix}
      \otimes
      \begin{bmatrix}
        1 \\[3pt]
        0
      \end{bmatrix}
    $
  }
  \caption{An illustration of pruning. Zeroing out alternating rows in a matrix is equivalent to a Kronecker product with the vector
  $\bigl[1\;0\bigr]^T$.}
  \label{fig:small-pruning-illustration}
\end{figure}

\section{Experiments and results}

\subsection{Training setup}

For pre-training, we follow general industry standards, namely those recommended by the Chinchilla paper \cite{hoffmann2022training}. 
We observed more stable loss curves with larger batch sizes. Consequently, we typically use a batch size of 120 sequences, attained with a gradient accumulation of $5$, with each forward pass seeing a batch of $24$ sequences, which amounts to approximately $122$k tokens per step. We use  a cosine scheduler for the \textbf{learning rate}, peaking at $6 \times 10^{-5}$ and diminishing to $6 \times 10^{-6}$, with a warm-up ranging from $500$ to $1500$ steps, depending on the model. We pre-train most our compressed models for 3 epochs, and use 4 epochs for the smaller models, on the OpenWebText corpus \cite{Gokaslan2019OpenWeb}. 

After pre-training, we use the best-performing checkpoint and continue training for one epoch with approximately $327$K tokens per step (320 sequences per step), with a very small constant learning rate, ideally $6 \times 10^{-6}$. 
All experiments were conducted on a single A100 80GB GPU.

\subsection{Model Naming, and Evaluation}
\label{sec:naming}

We adopt the nomenclature \textbf{Krony-PT-XM\{-Y-factors\}}, where $X$ denotes the parameter count and optionally $Y$ the number of Kronecker factors. Omitting the number of factors implicitly implies that only one factor was used. We focus on two model classes: \textbf{Krony-PT-81M} and \textbf{Krony-PT-96M}. To evaluate their performance, we measure next-token prediction perplexity on three standard benchmarks: wikitext103, wikitext2 \cite{merity2016pointer} and Lambada \cite{paperno2016lambada}. 

\subsection{The 81M class}
\label{sec:81M}

Table ~\ref{tab:perplexity-results} shows the results of our 81M model compared to other models of the same weight count category, namely DistilGPT2 \cite{sanh2019distilbert}, a distilled \cite{hinton2015distilling} version of GPT-2. Our model outperforms DistilGPT2 on all datasets, but especially on Lambada, which has a larger test set compared to both wikitext datasets. Perplexity scores of GPT-2 are also added for reference, all models significantly drop in performance, notably on Lambada.

\begin{table}[!t]
  \renewcommand{\arraystretch}{1.2}  
  \caption{Perplexity results of Krony-PT and DistilGPT2}
  \label{tab:perplexity-results}
  \centering
  \begin{tabular}{|c|l|c|c|c|}
    \hline
    \multicolumn{2}{c|}{\textbf{Model}} 
      & \multicolumn{3}{c|}{\textbf{Datasets}} \\ 
    \hline
    \textbf{\# Params} & \textbf{Model Name} 
      & \textbf{wikitext-103} & \textbf{wikitext-2} & \textbf{Lambada} \\ 
    \hline
    124~M & GPT-2 
      & 29.16 & 24.67 & 45.28 \\ 
    82~M  & DistilGPT2 
      & 44.53 & 36.48 & 76.00 \\ 
    82~M  & \textbf{KronyPT-81M} 
      & 42.74 & 35.75 & \textbf{61.02} \\ 
    80~M  & \textbf{KronyPT-80M-2} 
      & \textbf{41.06} & \textbf{34.56} & 66.88 \\ 
    \hline
  \end{tabular}
\end{table}

Table \ref{tab:perplexity-comparison-2} compares  \textbf{Krony-PT-81M} with larger Kronecker-based GPT-2 compressions. Section \ref{sec:param-count} details our parameter-counting method and how it differs from previous work. Although the 81-million-parameter variant lags behind the larger models on WikiText-103, it still outperforms all of them on LAMBADA. Note that the LAMBADA validation set we evaluate on ($\approx$ 400 k tokens) is considerably larger than that of WikiText-103 ($\approx$ 230 k tokens).

\begin{table}[!t]
  \renewcommand{\arraystretch}{1.2}  
  \caption{Perplexity of Krony-PT against other Kronecker-based models}
  \label{tab:perplexity-comparison-2}
  \centering
  \begin{tabular}{|c|l|c|c|c|}
    \hline
    \multicolumn{2}{|c|}{\textbf{Model}}
      & \multicolumn{3}{c|}{\textbf{Datasets}} \\
    \hline
    \textbf{\# Params} & \textbf{Model Name}
      & \textbf{wiki-103} & \textbf{wiki-2} & \textbf{Lambada} \\
    \hline
    81~M   & \textbf{KronyPT-81M}   & 42.74 & 35.75 & \textbf{61.02} \\ 
    119~M  & TQCompressedGPT2       & \textbf{40.28} & \textbf{32.25} & 64.72 \\
    119~M  & KnGPT-2                & 40.97 & 32.81 & 67.62 \\
    \hline
  \end{tabular}
\end{table}


\subsection{Pruning initialization compared to Van Loan:}

Table \ref{tab:prune-results} presents results for our proposed pruning-based initialization applied to \textbf{Krony-PT-96M} (denoted \textbf{Krony-PT-96M-prune}) alongside other Kronecker-compressed GPT-2 variants~\cite{tahaei2022kroneckerbert,abronin2024tqcompressor}. Our method consistently outperforms every competitor, including \textbf{Krony-PT-96M-VL}, which shares the same architecture but uses the conventional Van~Loan initialization. Our pruning based method is also faster to compute because it avoids the SVD step required by Van Loan. As in earlier experiments, the improvement is most pronounced on LAMBADA.

\begin{table}[!t]
  \renewcommand{\arraystretch}{1.2}  
  \caption{Results of pruning-based initialization}
  \label{tab:prune-results}
  \centering
  \begin{tabular}{|c|l|c|c|c|}
    \hline
    \multicolumn{2}{|c|}{\textbf{Model}}
      & \multicolumn{3}{c|}{\textbf{Datasets}} \\
    \hline
    \textbf{\# Params} & \textbf{Model Name}
      & \textbf{Lambada} & \textbf{wiki-103} & \textbf{wiki-2} \\
    \hline
    96~M  & \textbf{KronyPT-96M-prune}
         & \textbf{54.52} & \textbf{38.22} & \textbf{32.19} \\
    96~M  & \textbf{KronyPT-96M-VL}
         & 64.92          & 41.98          & 34.99         \\
    119~M & TQCompressedGPT2
         & 64.72          & 40.28          & 32.25         \\
    119~M & KnGPT-2
         & 67.62          & 40.97          & 32.81         \\
    \hline
  \end{tabular}
\end{table}

\subsection{Multiple Kronecker Factors with Normalized Scalars}
\label{multi-kron}

In this section, we study the effect of using multiple Kronecker factors, each multiplied by a scalar that is normalized as described in Section \ref{sec:norm-init}.
To keep the total parameter count comparable, we evaluate two settings:
\begin{enumerate}
    \item \textbf{Baseline (96M):} Single-factor model described in Section \ref{sec:96M}, initialized with the standard Van Loan (VL) decomposition.
    \item \textbf{Multiple factors setting (92.2M):} each MLP weight $W$ is initialized with a sum of $4$ Kronecker factors, each scaled by a learned scalar (i.e., $W  = \sum_{i=1}^{4} s_i (A_i \otimes B_i)$). The scalars $s_i$ follow our scaled VL initialization; each $A_i$ has size $(256,1024)$.
\end{enumerate}

Table \ref{tab:combined-table} reports the perplexity of both models.\footnote{Training hyper-parameters might not be fully optimal for the baseline; see Section \ref{sec:limitations}.}
The multiple-factor model with adaptive normalization consistently outperforms the baseline, achieving lower perplexity on every dataset after one epoch and retaining the lead after full training, with the largest gain on LAMBADA.

\begin{table}[!t]
  \renewcommand{\arraystretch}{1.2}
  \caption{Effect of adding multiple factors with normalized scalars after one epoch and full training.}
  \label{tab:combined-table}
  \centering
  \begin{tabular}{|l|c|c|c|c|}
    \hline
    \textbf{Model spec.}
      & \textbf{Training}
      & \textbf{Lambada}
      & \textbf{wiki-103}
      & \textbf{wiki-2} \\
    \hline
    Classic VL             & 1\,epoch        & 76.95       & 47.42       & 56.69       \\
    Adap. normalized VL & 1\,epoch        & \textbf{68.38} & \textbf{45.13} & \textbf{39.50} \\
    \hline
    Classic VL             & Full training   & 64.92       & 41.98       & 34.99       \\
    Adap. normalized VL & Full training   & \textbf{60.74} & \textbf{40.78} & \textbf{34.75} \\
    \hline
  \end{tabular}
\end{table}

Additionally, adding a scalar per Kronecker factor increases the parameter count only marginally. For instance, the 92.2M model introduced above adds a negligible $96$ scalars
\footnote{Depicted as follows: $96 = (\underbrace{4}_{\text{number of factors}} \times \underbrace{2}_{\text{MLP matrices per layer}}) \times \underbrace{12}_{\text{number of layers}}$} to the total parameter count.
The inference cost is also unchanged: after training, each scalar can be absorbed into $A_i$ (i.e., $A'_i = s_i A_i$), and serve the model for inference with each MLP weight represented as $W  = \sum_{i=1}^{4} A'_i \otimes B_i$.

\begin{figure}[!t]
  \centering
  \includegraphics[width=\columnwidth]{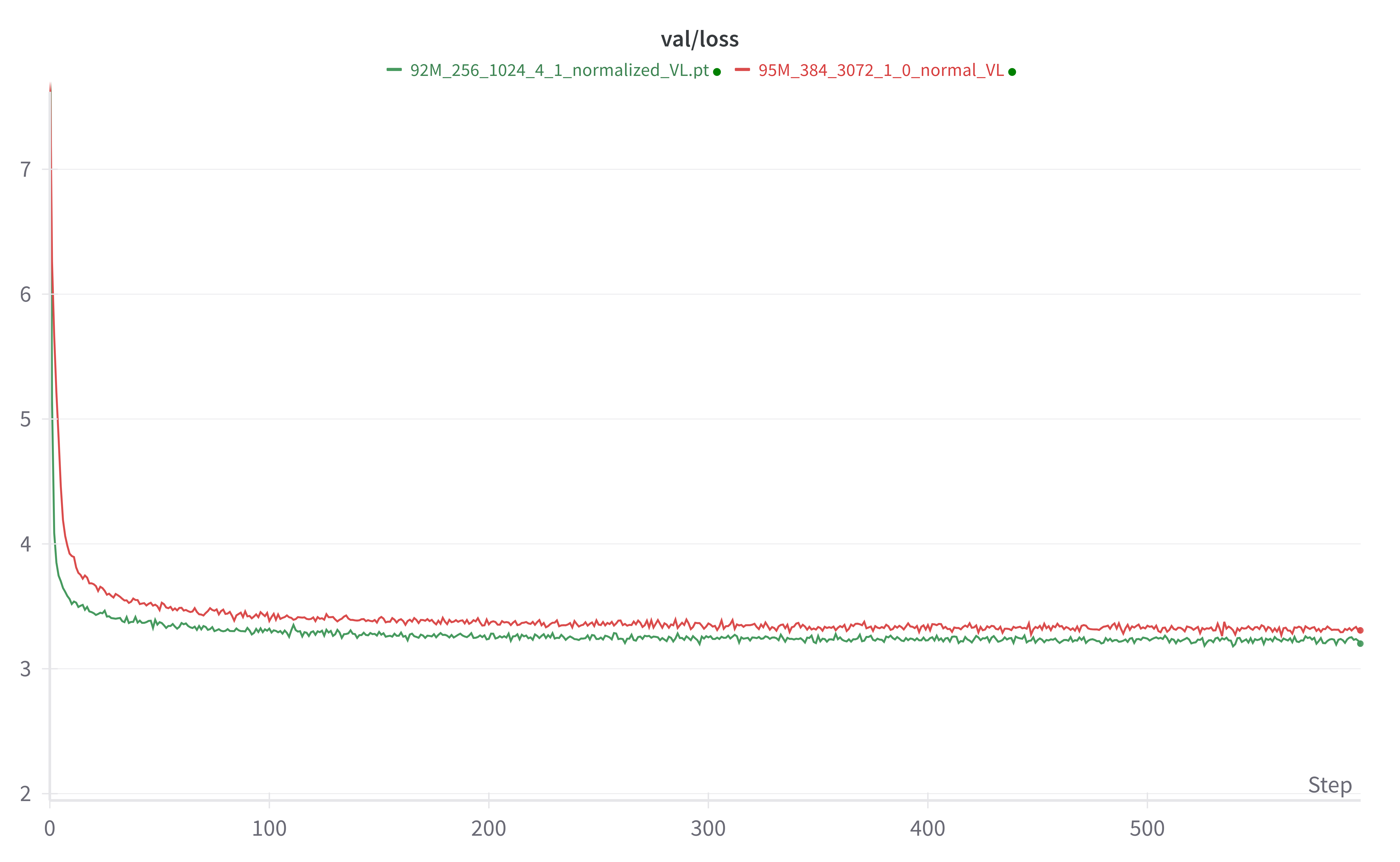}
  \caption{Negative log‐likelihood scores during the first 30\% of an epoch on the OpenWebText validation set.}
  \label{fig:negloglik-first-40pct}
\end{figure}

\textbf{Convergence speed.} Adding multiple Kronecker factors with adaptive, normalized scalars accelerates convergence, as shown in Table~\ref{tab:norms-helps-conv}. The 92\,M model with adaptive normalization achieves the baseline one-epoch perplexity after only processing 30\,\% of the first epoch and performs especially well on both WikiText datasets. 
Training curves show the same trend. The negative log-likelihood on the \textit{OpenWebText} validation set drops faster for the 92\,M model than for the 96\,M baseline with standard VL initialization.
Fig. ~\ref{fig:negloglik-first-40pct} shows the first 30\,\% of the epoch, highlighting this earlier convergence trend. This faster training translates into better generalization, as the same improvements appear in downstream benchmarks that are observed in Table~\ref{tab:norms-helps-conv}.

\begin{table}[!t]
  \renewcommand{\arraystretch}{1.2}
  \caption{Comparing perplexity during epochs.}
  \label{tab:norms-helps-conv}
  \centering
  \begin{tabular}{|l|c|c|c|c|}
    \hline
    \textbf{Model} & \textbf{Epoch \%} & \multicolumn{3}{c|}{\textbf{Datasets}} \\ 
    \cline{3-5}
                   &                   & \textbf{Lambada} & \textbf{wikitext-103} & \textbf{wikitext-2} \\ 
    \hline
    96\,M          & 100\% epoch       & 76.95            & 57.46                 & 47.42               \\ 
    \hline
    92\,M          & 30\% epoch        & \textbf{74.57}   & \textbf{45.19}        & \textbf{36.65}      \\ 
    \hline
  \end{tabular}
\end{table}

\section{Conclusion:}

In this work, we introduced Krony-PT, a compression technique for decoder-only models like GPT-2 using Kronecker Products. We systematically compress the feed-forward matrices of each transformer block, and explore different compression schemes resulting in models of various sizes. Our 81M model outperformed DistilGPT2 on next-token prediction and performed competitively with other compressed models of larger size. We also presented a new pruning-based initialization technique, and an improved scale-invariant Van Loan decomposition. We also used weight tying to further reduce model size. Our results highlight the potential of Kronecker-based compression to make large language models more efficient and accessible.

\section{Future work and limitations:}
\label{sec:limitations}

Future work includes investigating the impact of freezing scalar values at 4 when using four factors, as we observed that these scalars consistently converge to around 4 regardless of initialization. Additionally, we plan to explore inference speed improvements using advanced Kronecker Product tensor computations and further study the interpretability of each Kronecker factor. We also acknowledge the incomplete ablation studies due to limited computational resources, as we have not tested all training configurations for competing models.

\clearpage
\bibliographystyle{IEEEtran}
\bibliography{my-bib}

\clearpage

\appendices
\section{Kronecker Decomposition Algorithm}
\label{appendix:kronecker}

As discussed in Section~\ref{sec:VL}, the optimal Van Loan decomposition can be generated with the following Python script:

\begin{lstlisting}[language=Python]
def kronecker_decompose(A, m: int, n: int, *, k: int = 1, niter: int = 10):

    m2 = A.shape[-2] // m
    n2 = A.shape[-1] // n

    A = rearrange(A, "... (m m2) (n n2) -> ... (m n) (m2 n2)", m=m, m2=m2, n=n, n2=n2)

    u, s, v = torch.svd_lowrank(A, q=k, niter=niter)

    u = rearrange(u, "... (m n) k -> ... k m n", m=m, n=n, k=k)
    v = rearrange(v, "... (m2 n2) k -> ... k m2 n2", m2=m2, n2=n2, k=k)

    scale = s[..., None, None].sqrt()
    return u * scale, v * scale
\end{lstlisting}

\section{Parameter count}
\label{sec:param-count}
Other papers \cite{edalati2021kroneckr,abronin2024tqcompressor} report parameter counts differently from the convention we (and \emph{DistilGPT-2}) follow, making direct comparisons misleading. Their key deviation is to exclude the output embedding matrix, which is substantial (about 40 million parameters) when inherited from pre-trained GPT-2. As KnGPT2 \cite{edalati2021kroneckr} notes: ``the number of parameters of the models is reported excluding the output embedding layer in language modelling, which is not compressed, and is equal to row Parameters''. TQCompressor \cite{abronin2024tqcompressor} adopts the same practice, as clarified in a GitHub discussion\footnote{Link to GitHub issue: \url{https://github.com/terra-quantum-public/TQCompressedGPT2/issues/1}}. This difference means that their 81\,M-parameter model actually contains about 120\,M parameters. We do not count the output matrix because we use weight tying, as in the original GPT-2 paper.

\section{Only 3\% of data}

Both \cite{tahaei2022kroneckerbert} and \cite{abronin2024tqcompressor} state using only 3\% of the training data, this doesn’t reflect a stronger compression scheme, for the following reasons:

\begin{enumerate}
	\item They only factorize the odd‐indexed transformer blocks, hence, over half the parameters of the compressed model fully match the fully pretrained checkpoint of GPT-2.
	\item Within the odd layers, the Kronecker factors are not randomly initialized, as the Van Loan SVD-based decomposition is used which retains the top singular components (and thus the core pretrained features).
	\item Most importantly, they reuse GPT-2’s original 38 M-parameter output projection unchanged—and they do so without any weight tying—meaning the entire pretrained decoder vocabulary mapping is carried over intact.
\end{enumerate}

Consequently, the precise amount of pretrained knowledge retained is unknown. A fair baseline would randomly initialize all compressed-model parameters, without reusing any weights from the original checkpoint, which is not the case here.

\end{document}